\documentclass[sigconf]{acmart}
\pdfoutput=1

\usepackage{makecell}
\usepackage{multirow}
\usepackage{balance}
\usepackage{bm}

\AtBeginDocument{%
  \providecommand\BibTeX{{%
    \normalfont B\kern-0.5em{\scshape i\kern-0.25em b}\kern-0.8em\TeX}}}

\copyrightyear{2022}
\acmYear{2022}
\setcopyright{acmcopyright}
\acmConference[MM '22] {Proceedings of the 30th ACM International Conference on Multimedia }{October 10--14, 2022}{Lisbon, Portugal.}
\acmBooktitle{Proceedings of the 30th ACM International Conference on Multimedia (MM '22), October 10--14, 2022, Lisbon, Portugal}
\acmPrice{15.00}
\acmISBN{978-1-4503-9203-7/22/10}
\acmDOI{10.1145/3503161.3547916}

\acmSubmissionID{mmfp0705}

\begin{document}

\title{UDoc-GAN: Unpaired Document Illumination Correction with Background Light Prior}

\author{Yonghui Wang$^1$, \quad Wengang Zhou$^{1,2}$$^{\dagger}$, \quad Zhenbo Lu$^2$, \quad Houqiang Li$^{1,2}$$^{\dagger}$}


\affiliation{%
 $^1$\institution{CAS Key Laboratory of Technology in GIPAS, EEIS Department, University of Science and Technology of China}
 $^2$\institution{Institute of Artificial Intelligence, Hefei Comprehensive National Science Center} 
 \country{}
}
\email{wyh1998@mail.ustc.edu.cn, zhwg@ustc.edu.cn, luzhenbo@iai.ustc.edu.cn, lihq@ustc.edu.cn}
\thanks{$^{\dagger}$Corresponding authors: Wengang Zhou and Houqiang Li.}
\renewcommand{\shortauthors}{Yonghui Wang, et al.}
\begin{abstract}
Document images captured by mobile devices are usually degraded by uncontrollable illumination, which hampers the clarity of document content. 
Recently, a series of research efforts have been devoted to correcting the uneven document illumination.
However, existing methods rarely consider the use of ambient light information, and usually rely on paired samples including degraded and the corrected ground-truth images which are not always accessible.
To this end, we propose UDoc-GAN, the first framework to address the problem of document illumination correction under the unpaired setting. 
Specifically, we first predict the ambient light features of the document.
Then, according to the characteristics of different level of ambient lights, we re-formulate the cycle consistency constraint to learn the underlying relationship between normal and abnormal illumination domains. 
To prove the effectiveness of our approach, we conduct extensive experiments on DocProj dataset under the unpaired setting. 
Compared with the state-of-the-art approaches, our method demonstrates promising performance in terms of character error rate (CER) and edit distance (ED), together with better qualitative results for textual detail preservation.
The source code is now publicly available at \url{https://github.com/harrytea/UDoc-GAN}.

\end{abstract}

\ccsdesc[500]{Computing methodologies~Computer vision}
\ccsdesc[500]{Networks~Network design principles}


\keywords{Document Illumination Correction, Unpaired Image-to-Image Translation, Cycle Consistency, OCR}


\begin{teaserfigure}
  \includegraphics[width=\textwidth]{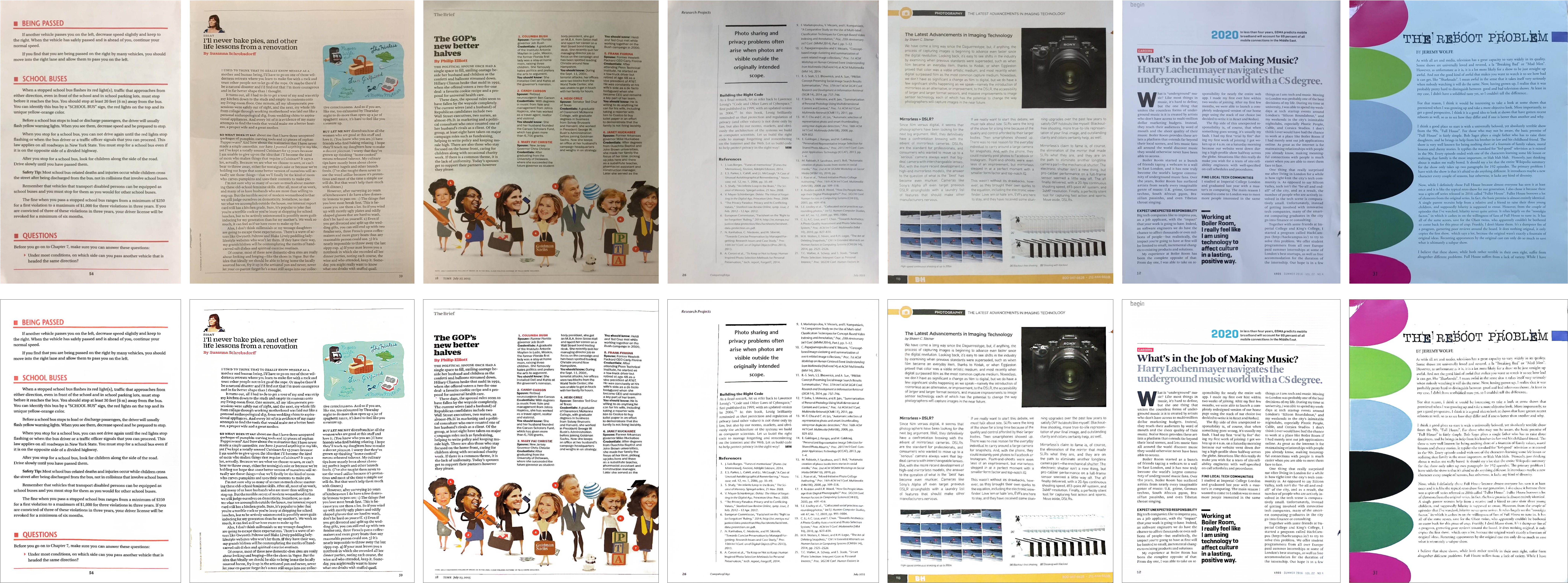}
  \caption{Qualitative results of our proposed UDoc-GAN. The top row shows the geometric correction results of DocTr~\cite{feng2021doctr}. The second row presents the illumination correction results of our approach. }
  \label{fig:teaser}
\end{teaserfigure}

\maketitle

\section{Introduction\label{sec:intro}}

\begin{figure}[tbp]
	\begin{center}
		\includegraphics[width=0.98\linewidth]{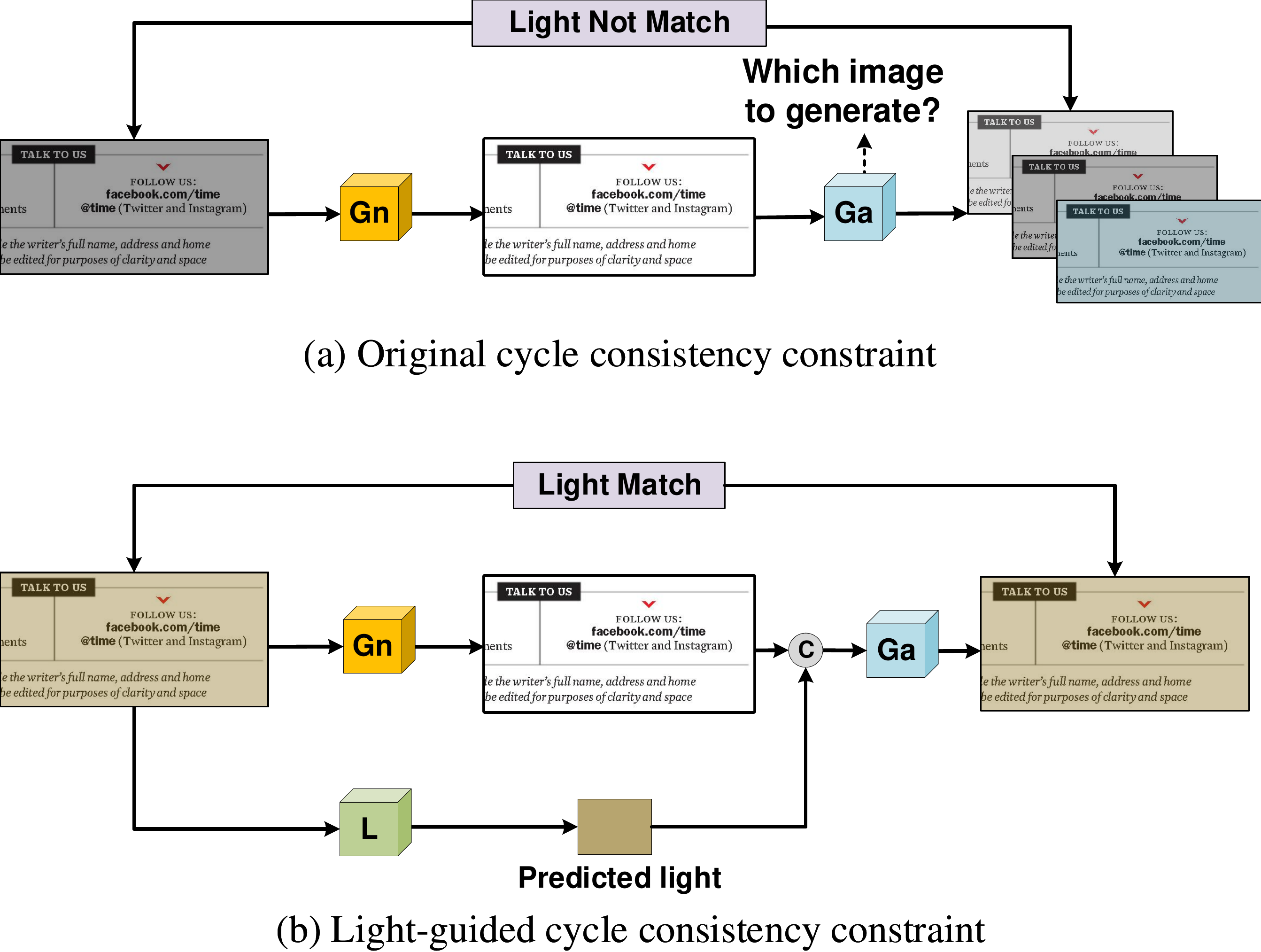}
	\end{center}
	\vspace{-0.1in}
	\caption{The architecture of cycle consistency constraint. (a) Original form. \bm{$G_{a}$} generates images without any guidance. (b) Light-guided form. Guided by the light prediction network \bm{$L$}, \bm{$G_{a}$} can generate a specific illumination image to suffice the cycle consistency constraint.}
	\label{cycle}
\end{figure}

The rapid development of digital cameras has led to a surge of applications that digitize document files by directly taking photos with mobile devices.
Different from document scanning using the traditional scanner, the document images taken by handheld mobile devices usually suffer from uncontrollable shooting angles and complex ambient light, resulting in uneven illumination and color deviation. 
The unsatisfied visual quality not only affects the acquisition of visual information by human eyes, but also poses a negative impact on the subsequent processing tasks, such as Optical Character Recognition (OCR). 
Therefore, document illumination correction has become an important task in intelligent document processing systems and attracts increasing research interests~\cite{9010747,li2019document,feng2021doctr,9156786}. 

At the early stage, illumination correction methods mainly adopt the concept of intrinsic images,  \emph{e.g.}, light contrast~\cite{brown2006geometric,zhang2007photometric} and albedo distribution~\cite{DBLP:journals/ijcv/WadaUM97}. 
With the successful application of deep learning techniques in a variety of computer vision tasks, recent methods resort to deep neural networks for document illumination correction ~\cite{9010747,li2019document,feng2021doctr}. 
These methods generally follow two paradigms to address the problem of illumination correction. One paradigm is based on network architecture design, the other paradigm is to leverage additional information, such as shading maps.

Despite achieving significant improvements, the current state-of-the-art methods~\cite{9010747,li2019document,feng2021doctr} commonly ignore the function of light prior, and only rely on a data-driven manner.
Although these methods can partially recover the semantic information in the document, they are not friendly to the recovery of texture and color. 
Intuitively, when a document image is affected by ambient light, the color texture of the document will change. 
Specifically, for the background of the document, it will be converted to a color similar to the background light. 
But for the text and pictures in the document, after being affected by the ambient light, there will be different color texture deviations in different regions. 
Therefore, the illumination correction of the document should not only restore the content information of the document but also restore the original texture and color of the text and pictures. 
In addition, existing methods heavily rely on paired datasets including abnormal and corresponding normal illumination images, which brings extra burdens to the task. 
By applying the light prior, we can solve the document illumination correction problem under an unpaired setting.

Given a captured digital image, we assume that the document is an ideal diffuse reflector, even if the overall illumination is uneven, but most regions exhibit uniform illumination.
Based on this hypothesis, we propose a generative adversarial network, UDoc-GAN, which takes the background light as a prior to correct illumination issues. 
As shown in Fig.~\ref{cycle}(a), we aim to train a normal illumination image generator $G_{n}$, which takes an abnormal illumination image as input, and generates an image which is indistinguishable from the normal illumination image. But only applying this setting, the mapping is under-constrained. So we train another network $G_{a}$, which generates abnormal illumination images, to learn the inverse mapping. However, the mapping between the normal and abnormal illumination domains is not one-to-one (\emph{i.e.}, the generator $G_{a}$ is unaware of what kind of illumination image to generate). 
Then, we propose a light prediction network $L$ to deal with this problem. As shown in Fig.~\ref{cycle}(b), from an input abnormal illumination image, we produce a normal illumination image, and predict what kind of ambient light the input image is suffered from. Then, $G_{a}$ can produce an output which matches the corresponding input.
In such a way, we can employ the unpaired training via the light-guided cycle consistency constraint.
Experiment results demonstrate that our method achieves superior performance compared with the state-of-the-art methods on the existing benchmark datasets.


In conclusion, we make three-fold contributions:
\begin{itemize}
\item We propose a novel UDoc-GAN framework to address the problem of document illumination correction. Different from the previous methods relying on paired samples, our proposed UDoc-GAN is optimized under an unpaired setting. To the best of our knowledge, this is the first method using unpaired data for document illumination correction.

\item Our approach outperforms the state-of-the-art methods in most indicators on the existing benchmark~\cite{ma2018docunet}, even using an unpaired dataset. 
Meanwhile, the qualitative results also validate that the images corrected by our method show more consistent color with the ground truth image. 

\item Compared with other state-of-the-art methods, our method is more computationally efficient at the inference stage. Unlike existing methods that crop each patch for pre-processing, our proposed network can directly load the whole document image as input, which makes the proposed method more applicable. 
\end{itemize}

\section{RELATED WORK}

\subsection{Document Illumination Correction}
Document illumination correction stems from the document rectification problem~\cite{937649, zhang2008improved, meng2014active}, which has been investigated for many years. Early works mainly adopt the concept of intrinsic images to correct the uneven illumination of documents. 
Wada \emph{et al}.\cite{DBLP:journals/ijcv/WadaUM97} employs the illumination information to conduct the geometric correction and slightly mitigates the uneven illumination effect. Brown \emph{et al}.~\cite{brown2006geometric} and Zhang \emph{et al}.~\cite{zhang2007photometric} solve the uneven illumination problem by separating illuminations from the original image. Some researches\cite{bako2016removing, jung2018water, 9156786} design various algorithms to alleviate the influence of shadow regions in documents. 

Recently, several studies suggest that deep learning is beneficial to the uneven illumination problem. Li \emph{et al}.~\cite{li2019document} use residual blocks~\cite{he2016deep} with skip connections to learn the luminance residual. 
Das \emph{et al}.~\cite{9010747} design a network which consists of two U-Net style encoder-decoders. One is used to predict surface normal, and the other estimates shading maps, which jointly help to solve the uneven illumination problem. 
Lin \emph{et al}.~\cite{9156786} design a network to remove the shadow for document images. They also utilize an extra network to predict the background color and generate the attention map for shadow removal. Feng \emph{et al}.~\cite{feng2021doctr} propose a transformer-based method to address the issue of geometry and illumination distortion of document images and can achieve good performance in quantitative results. These studies collectively indicate that the data-driven methods shed light on a promising way to document illumination correction problem.

\subsection{Generative Adversarial Networks}
GANs~\cite{goodfellow2014generative} have been widely used in various generative tasks~\cite{zhu2016generative, mathieu2016disentangling, yu2017unsupervised} since it was invented. In addition, GANs have been adopted in many other applications such as object detection~\cite{li2017perceptual, bai2018sod, prakash2021gan}, image super-resolution~\cite{ledig2017photo, guan2019srdgan} and medical image processing~\cite{xue2018segan, choi2017generating}. The essence of GANs can be summarized as training of two networks simultaneously. One is the generator network and the other is the discriminator network. Moreover, the primary loss for GANs is adversarial loss, which makes the generator and discriminator resist each other, aiming to generate images that the discriminator can not distinguish from the real ones. Although GANs can accomplish such challenging tasks, the training of GANs is unstable and often has various problems, such as mode collapse~\cite{arora2017generalization}, vanishing gradient~\cite{goodfellow2016nips}, and internal covariate shift~\cite{ioffe2015batch}.

\subsection{Cycle Consistency}
Cycle consistency constraint has been used in many fields~\cite{zach2010disambiguating, huang2013consistent, wang2013image, he2016dual}. Recent works~\cite{godard2017unsupervised, zhou2016learning} use cycle consistency loss as a constraint to supervise the network training. CycleGAN~\cite{zhu2017unpaired}, which is the most similar to our method, uses two generative adversarial networks to formulate cycle consistency constraints. Any one of them can transfer one domain from another. Namely, the network translates an image from domain $X$ to domain $Y$ and then translates it back (\emph{i.e.}, $X\rightarrow Y\rightarrow X$). The ultimate image should be the same as the input. However, without the cycle consistency constraint (\emph{i.e.}, only $X\rightarrow Y$), the network may choose an easy way to map many images into one image so as to simplify training which is called mode collapse. By enforcing the cycle consistency mechanism, the CycleGAN architecture can prevent generator from extreme hallucinations~\cite{jabbar2021survey}. Consequently, cycle consistency constraint plays a crucial role in avoiding mode collapse problem and makes the training of GANs more stable~\cite{zhu2017unpaired}.

\begin{figure}[tbp]
	\begin{center}
		\includegraphics[width=0.98\linewidth]{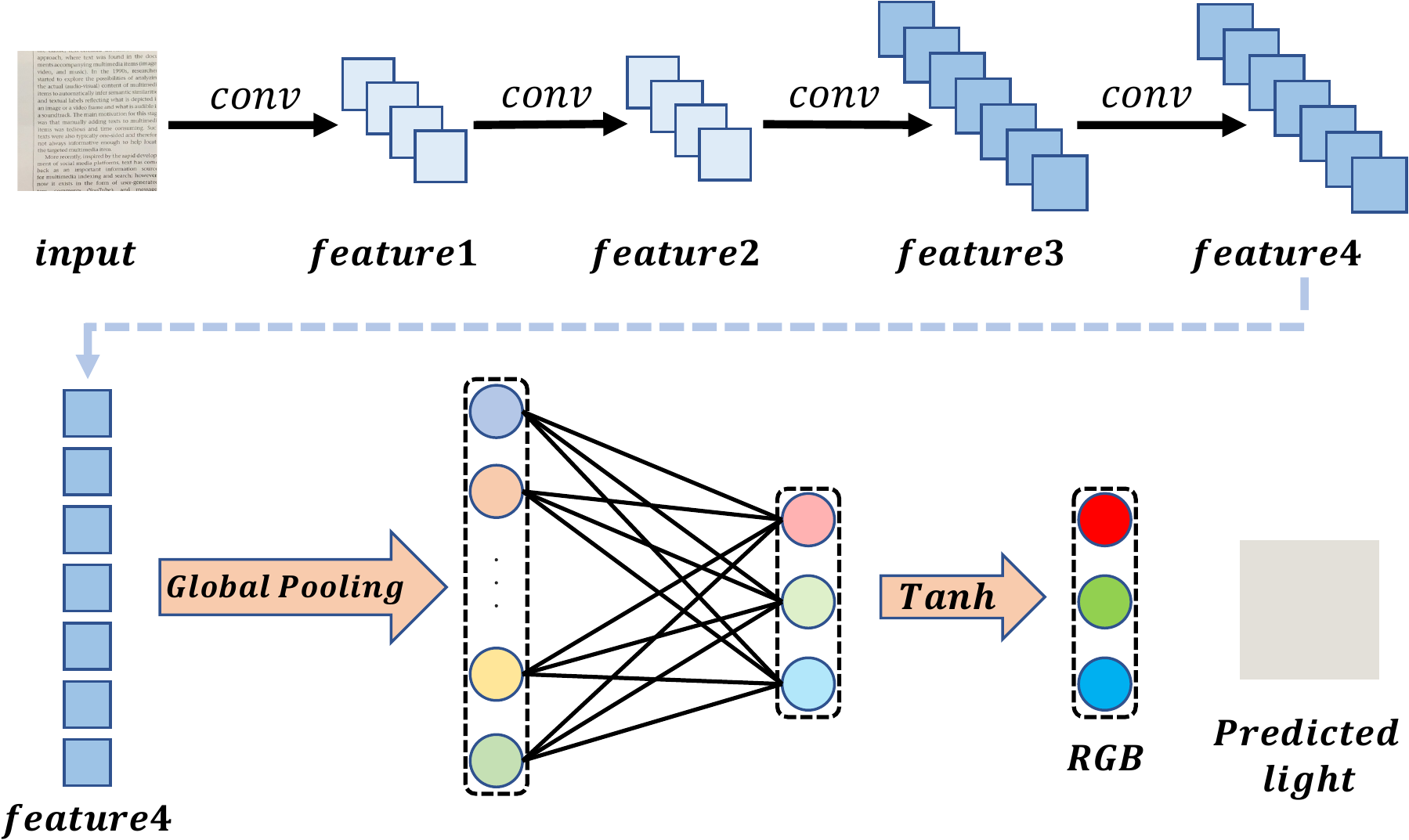}
	\end{center}
	\vspace{-0.1in}
	\caption{An illustration of Light Prediction Network (LPNet). LPNet first exploits four convolutional layers to extract spatial information of the input image, then adopts global max-pooling operation followed by a fully connected layer to generate a 3-dim global color statistic vector. Finally, a \textit{Tanh} activation function is applied to standardize the output.}
	\label{fig:LPNet}
\end{figure}


\begin{figure*}[tbp]
	\begin{center}
		\includegraphics[width=0.98\linewidth]{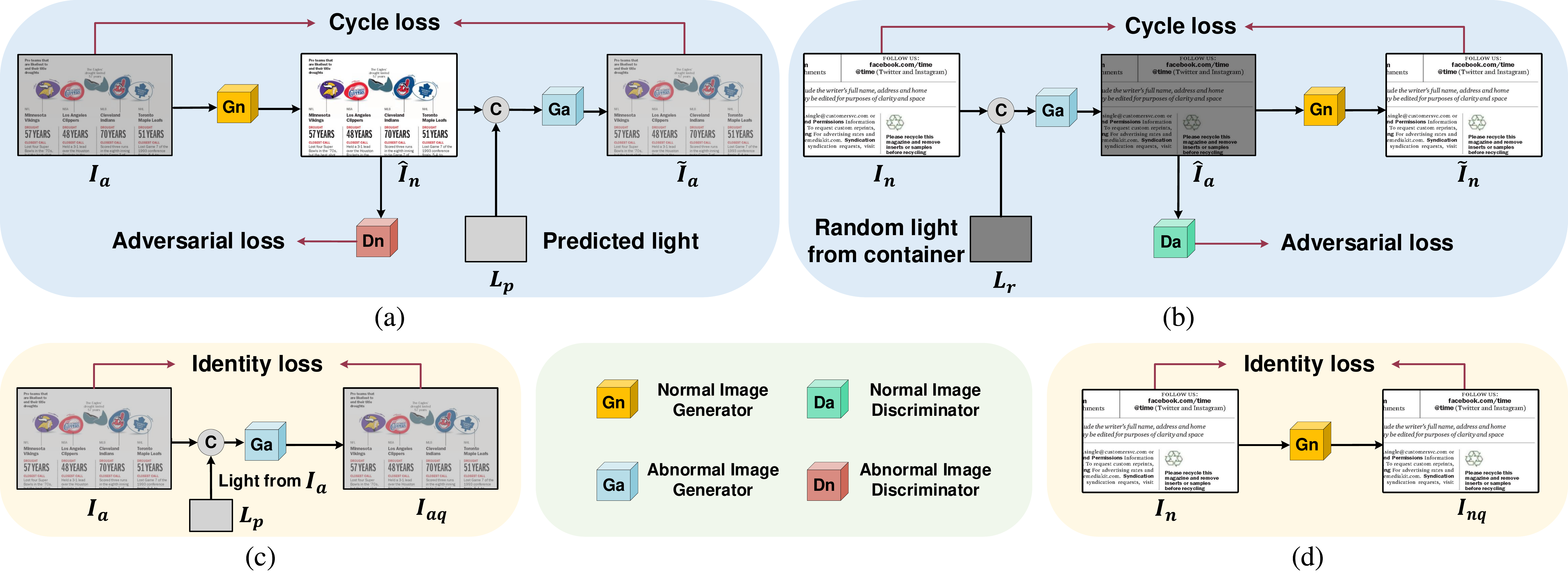}
	\end{center}
	\vspace{-0.1in}
	\caption{Overall architecture of our proposed UDoc-GAN, which has two parts: (a) and (c) illustrate the abnormal illumination images generation pipeline with the light prior. (b) and (d) illustrate the normal illumination images generation pipeline with the light prior. Each part includes three losses: cycle consistency loss, adversarial loss and identity loss. Additionally, $\bm{I}_{a}$ and $\bm{I}_{n}$ denote the abnormal and normal illumination images. $\tilde{\bm{I}}_{a}$ and $\bm{I}_{aq}$ are the generated abnormal illumination images while $\tilde{\bm{I}}_{n}$ and $\bm{I}_{nq}$ are the generated normal illumination images. $\bm{L}_{p}$ and $\bm{L}_{r}$ are the light priors.}
	\label{fig:overall}
\end{figure*}

\section{Our Approach}

In this section, we elaborate on the overall architecture of our proposed method, which consists of two parts: light prediction network and unpaired document illumination correction network.

\subsection{Light Prediction Network\label{sec:lpnet}}

As noted above, to formulate the light-guided cycle consistency constraint, the foremost thing to do is to estimate the background light source. To do this, we make some modifications to DocProj dataset~\cite{li2019document}. For each document image, due to the influence of lights and shadows, the background color is various, such as grey, yellow, etc. Considering that the illumination of the document is generally uneven, we crop each sample in the dataset and only select the uniformly illuminated part. Then, we manually extract the background color of each part as its ground truth light. We try to select as many samples of illumination conditions as possible, in order to train LPNet to ensure that it can make good predictions of illuminations and has strong generalization performance.

The light prediction network is shown in Fig.~\ref{fig:LPNet}. In the top row, given an image $\bm{I}_{d} \in \mathbb{R}^{H \times W \times 3}$, which is affected by particular light, we first handle it with four Conv layers $H_{c_{i}}$, where $i = 0,1,2,3$. The output of each layer is $\bm{I}_{d}^{c_{i}} \in \mathbb{R}^{H \times W \times C_{i}}$, and we set $C_{i}=32, 32, 128, 128$, respectively. To dramatically accelerate the training speed as well as boost the network performance, batch normalization~\cite{ioffe2015batch} and ReLU activation function are added after every Conv layer. This process can be expressed as follows:
\begin{equation}
  \bm{I}_{m} = \delta_{r}(H_{b4}(H_{c4}(\cdot \cdot \cdot \delta_{r}(H_{b1}(H_{c1}(\bm{I}_{d})))\cdot \cdot \cdot))),
\end{equation}
where $H_{ci}(\cdot)$ denotes the $i^{th}$ Conv layer, $\bm{I}_{d}$ and $\bm{I}_{m}$ are input and output for the holistic Conv operations. $H_{bi}(\cdot)$ and $\delta_{r}(\cdot)$ denote batch normalization and ReLU activation function, respectively. After that, in the bottom row of Fig.~\ref{fig:LPNet}, we adopt global max pooling followed by a fully connected layer and finally activated by Tanh activation function. Global max pooling acts as a structural regularizer that prevents overfitting. Moreover, such global statistics can also be considered as descriptors of all feature maps, which will conduce to the prediction of background color. 
\begin{equation}
  \bm{I}_{RGB} = \delta_{t}(W_{l}G_{pool}(\bm{I}_{m})),
\end{equation}
where $\bm{I}_{RGB}$ is a $\mathbb{R}^{1 \times 3}$ vector containing the values of RGB channels. $G_{pool}(\cdot)$ denotes the global maxpooling function, $W_{l}$ and $\delta_{t}(\cdot)$ denote the weight of fully connected layer and Tanh activation function, respectively.

\subsection{Unpaired Illumination Correction}

Fig.~\ref{fig:overall} illustrates the overall architecture of our proposed network, UDoc-GAN, which intends to transfer domains between normal and abnormal illumination documents. 
Here, we will introduce our network from the perspective of how the light prior acts on the two domains. 
Before that, we will first introduce the light container.

\textbf{Light container.} Since we predict multiple background lights from abnormal illumination images, we adopt a container to conserve them. As for abnormal images, we have the predicted light stored in the container and then utilize it as a prior at the cycle consistency stage. Besides, we also use this light to regularize the abnormal image generator to output an image that is close to the input abnormal image. And for normal images, we randomly select a light in the container to conduct another cycle consistency stage. Empirically, we set the length of light container as a quarter of the number of samples in the dataset. When the container is full, we eliminate the primary light to provide space for the new light that the proceeding is regarded as \emph{a queue}.

\textbf{Prior for abnormal illumination documents.} As shown in Fig.~\ref{fig:overall}(a) and Fig.~\ref{fig:overall}(c), we first have an abnormal illumination document $\bm{I}_{a}$ be processed by generator $G_{n}$ and get a normal illumination document $\hat{\bm{I}}_{n}$. Then a discriminator $D_{n}$ is applied to differentiate whether $\hat{\bm{I}}_{n}$ is a real normal illumination document or not. Here, we get our first adversarial loss function as follows:
\begin{equation}
\begin{aligned}
    \mathcal{L}_{GAN}^{a}(G_{n}, D_{n}) &= \mathbb{E}_{\bm{I}_{n} \sim p_{data}(\bm{I}_{n})}[\log(D_{n}(\bm{I}_{n}))] \\
            & + \mathbb{E}_{\bm{I}_{a} \sim p_{data}(\bm{I}_{a})}[\log(1-D_{n}(G_{n}(\bm{I}_{a})))],
\end{aligned}
\end{equation}
where $\mathbb{E}$ and $p_{data}$ denote the error and the data distribution, respectively. $\bm{I}_{n} \sim p_{data}(\bm{I}_{n})$ and $\bm{I}_{a} \sim p_{data}(\bm{I}_{a})$ indicate that $\bm{I}_{n}$ and $\bm{I}_{a}$ are selected, respectively from the data distribution $p_{data}$ over the normal and abnormal illumination document datasets. $G_{n}$ aims to generate document $G_{n}(\bm{I}_{a})$ which looks like normal domain, and the purpose of $D_{n}$ is to determine whether $G_{n}(\bm{I}_{a})$ is a normal illumination image or not. For $D_{n}$, it aims to maximize the loss, while for $G_{n}$, it aims to minimize the above loss, \emph{i.e.}, $min_{G_{n}}max_{D_{n}}\mathcal{L}_{GAN}^{a}(G_{n}, D_{n})$. The existence of adversarial loss can assist the mutual transformation between normal and abnormal domains. However, only relying on adversarial loss to train the model may result in some artifacts on the generated images~\cite{isola2017image}. Moreover, with the powerful network, $G_{n}$ may choose a relatively simple way for mappings, such as mapping the documents in the abnormal domain to a certain or a few documents in the normal domain. This will change the content of the image, which severely deviates from the task objective of retaining the content information in the images and only changing its illumination information. Therefore, we add the second loss, cycle consistency loss, to preserve the information constantly. The cycle consistency constraint can be described as follows:
\begin{equation}
    \bm{I}_{a} \rightarrow G_{n}(\bm{I}_{a}) \rightarrow G_{a}(G_{n}(\bm{I}_{a})) \approx \bm{I}_{a},
\end{equation}
with respect to $G_{a}(\hat{\bm{I}}_{n}) \rightarrow \tilde{\bm{I}}_{a}$, the mapping is ill-posed because $G_{a}$ can map $\hat{\bm{I}}_{n}$ to various illumination conditions, which directly causes the mapping mismatch. Thus, to encourage the generated document $\tilde{\bm{I}}_{a}$ has uniform content and the same illumination with $\bm{I}_{a}$, we need to use the light container discussed in the previous part, which contains the light condition obtained from $\bm{I}_{a}$. By using the light $\bm{L}_{p}$ as prior, it can guide $G_{a}$ to generate image $\tilde{\bm{I}}_{a}$ which is similar to $\bm{I}_{a}$. Then it will suffice the cycle consistency constraint. This light-guided cycle consistency loss can be expressed as
\begin{equation}
    \mathcal{L}_{cycle}^{a}(G_{n}, G_{a}) = \mathbb{E}_{\bm{I}_{a} \sim p_{data}(\bm{I}_{a})}[\left \| G_{a}(G_{n}(\bm{I}_{a}), \bm{L}_{p})-\bm{I}_{a} \right \|_{1}].
\end{equation}
To make the training more stable, we use the $L_{1}$ norm as our loss to capture the difference between $\bm{I}_{a}$ and the restored document $\tilde{\bm{I}}_{a}$. $G_{a}$, which generates the abnormal illumination images, should keep the light constant when it is fed an abnormal light document image. Besides, we also add the light $\bm{L}_{p}$ from $\bm{I}_{a}$ to supervise the identity process, which is formulated as an identity loss. It preserves the original light of the input image:
\begin{equation}
    \mathcal{L}_{identity}^{a}(G_{a}) = \mathbb{E}_{\bm{I}_{a} \sim p_{data}(\bm{I}_{a})}[\left \| G_{a}(\bm{I}_{a}, \bm{L}_{p})-\bm{I}_{aq} \right \|_{1}],
\end{equation}
where $\bm{I}_{aq}$ is the reconstructed image guided by $\bm{L}_{p}$.

\begin{figure*}[th]
	\begin{center}
		\includegraphics[width=0.88\linewidth]{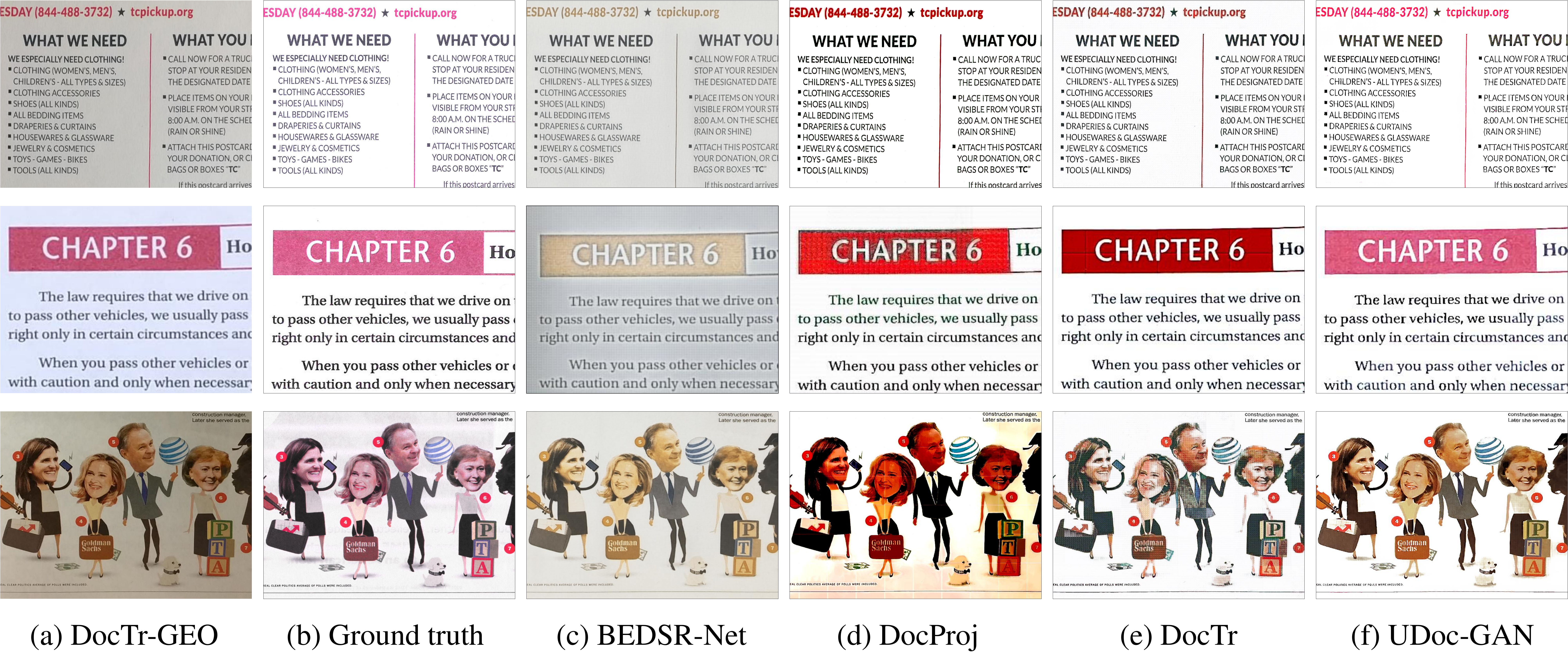}
	\end{center}
	\vspace{-0.1in}
	\caption{Qualitative comparisons between our proposed UDoc-GAN and other state-of-the-art methods. The illumination correction is performed based on the geometric rectification results of DocTr. }
	\label{fig:doctr}
\end{figure*}

\textbf{Prior for normal illumination documents.} Fig.~\ref{fig:overall}(b) and Fig.~\ref{fig:overall}(d) shows how prior knowledge is used in the normal domain. Starting from a normal illumination document $\bm{I}_{n}$, we take a random light $\bm{L}_{r}$ from the light container and use it to guide $G_{a}$ to generate an abnormal illumination image $\hat{\bm{I}}_{a}$. This image is used to fool the discriminator and makes $D_{a}$ challenging to distinguish if the image is real or fake. We express this adversarial loss as
\begin{equation}
\begin{aligned}
    \mathcal{L}_{GAN}^{b}(G_{a}, D_{a}) &= \mathbb{E}_{\bm{I}_{a} \sim p_{data}(\bm{I}_{a})}[\log D_{a}(\bm{I}_{a})] \\
            & + \mathbb{E}_{\bm{I}_{n} \sim p_{data}(\bm{I}_{n})}[\log(1-D_{a}(G_{a}(\bm{I}_{n}, \bm{L}_{r})))].
\end{aligned}
\end{equation}
In order to keep all kinds of information (\emph{i.e.}, color texture and text) in the images, we still utilize the cycle consistency constraint. Briefly, we formulate the objective as
\begin{equation}
    \bm{I}_{n} \rightarrow G_{a}(\bm{I}_{n}, \bm{L}_{r}) \rightarrow G_{n}(G_{a}(\bm{I}_{n}, \bm{L}_{r})) \approx \bm{I}_{n}.
\end{equation}
The corresponding loss function can be expressed as
\begin{equation}
    \mathcal{L}_{cycle}^{b}(G_{a}, G_{n}) = \mathbb{E}_{\bm{I}_{n} \sim p_{data}(\bm{I}_{n})}[\left \| G_{n}(G_{a}(\bm{I}_{n}, \bm{L}_{r}))-\bm{I}_{n} \right \|_{1}].
\end{equation}
Finally, to further enhance the performance of $G_{n}$ and reduce the possibility of color skew, we add identity loss which is the same as the previous step:
\begin{equation}
    \mathcal{L}_{identity}^{b}(G_{n}) = \mathbb{E}_{\bm{I}_{n} \sim p_{data}(\bm{I}_{n})}[\left \| G_{n}(\bm{I}_{n})-\bm{I}_{nq} \right \|_{1}],
\end{equation}
where $\bm{I}_{nq}$ is the reconstructed image which is identity with $\bm{I}_{n}$. In our experiments, the color texture of text and pictures may change if we do not add this loss.

\subsection{Loss Function}
In this section, we summarize our loss function of two proposed networks, LPNet and UDoc-GAN, respectively.

\textbf{LPNet.} We leverage the $L_{1}$ loss to minimize the error of LPNet, which can be described as follows:
\begin{equation}
    \mathcal{L}_{light} = \left \| \bm{I}_{RGB}-\bm{I}_{GT} \right \|_{1},
\end{equation}
where $\bm{I}_{RGB}$ is the predicted light and $\bm{I}_{GT}$ is the ground truth light of image, respectively.

\textbf{UDoc-GAN.} In summary, the loss of UDoc-GAN consists of there parts: adversarial loss, cycle consistency loss, and identity loss. The full objective can be formulated as
\begin{equation}
\begin{aligned}
    \mathcal{L}_{all}(G_{a}, G_{n}, D_{a}, D_{n}) &= \mathcal{L}_{GAN}^{a}(G_{a}, D_{a}) + \mathcal{L}_{GAN}^{b}(G_{n}, D_{n}) \\
            &+ \lambda_{1}(\mathcal{L}_{cycle}^{a}(G_{a}, G_{n}) + \mathcal{L}_{cycle}^{b}(G_{a}, G_{n})) \\
            &+ \lambda_{2}(\mathcal{L}_{identity}^{a}(G_{a}) + \mathcal{L}_{identity}^{b}(G_{n})),
\end{aligned}
\end{equation}
where $\lambda_{i}$ represents the weight of each item in the total loss. We follow~\cite{zhu2017unpaired, hu2019mask} and set $\lambda_{1}$ and $\lambda_{2}$ as 10 and 5, respectively. Briefly, we are committed to solving the problem:
\begin{equation}
    \arg \min\limits_{G_{a}, G_{n}} \max\limits_{D_{a}, D_{n}}\mathcal{L}_{all} (G_{a}, G_{n}, D_{a}, D_{n}).
\end{equation}

\section{EXPERIMENTS}
\subsection{Datasets}
Following~\cite{feng2021doctr}, we adopt DocProj dataset~\cite{li2019document} as our training dataset. Concretely, for LPNet, as mentioned in Sec.~\ref{sec:lpnet}, we cut out the region with uniform illumination from each sample of the dataset, and manually extract the background color as ground truth light. For UDoc-GAN, we randomly shuffle the dataset so that the normal and abnormal illumination images are not paired. During testing, we adopt DocUNet Benchmark~\cite{ma2018docunet}, which is also suggested by previous methods~\cite{ma2018docunet, 9010747, li2019document, feng2021doctr}, to evaluate the effectiveness of our method. The details of the two datasets are as follows.

\textbf{DocProj.} DocProj dataset~\cite{li2019document} is a synthetic dataset which consists of 2700 images. Each sample has a high resolution of $2400 \times 1800$. Considering that the dataset itself has geometric distortion, the first thing we do is to correct it with the geometric correction method provided by the original paper, and then we apply the dataset of geometric correction to carry out our subsequent experiments.

\textbf{DocUNet Benchmark.} DocUNet Benchmark~\cite{ma2018docunet} is a real world
dataset is taken with a mobile camera, which contains 130 images of various content and format. This benchmark is also geometrically distorted and therefore we use two geometric correction methods, DewarpNet~\cite{9010747} and DocTr~\cite{feng2021doctr}, to carry out the geometric correction, that is, there are only a few geometric distortions and the documents are only influenced by uneven illumination and a few creases.

\subsection{Evaluation Metrics}
The principal purpose of document illumination correction is to improve the perception effect and strengthen the character recognition accuracy. We use the image similarity metric, Multi-Scale Structural Similarity (MS-SSIM) to evaluate the perception effect. For a fair comparison, all results are resized to a 598400-pixel area as~\cite{ma2018docunet,9010747} do. Optical Character Recognition (OCR) is a technique that aims to convert printed or handwritten text into machine-encoded text and is subject to a multiplicity of factors, such as uneven illumination. For OCR, Edit Distance (ED) and Character Error Rate (CER) are the two most important indicators. We find that different methods select different images for testing, which may bring contingency. Thus we combine the images selected by~\cite{9010747} and~\cite{feng2021doctr} for OCR evaluation. To conclude, we use MS-SSIM to evaluate the perceptual quality, and ED and CER to evaluate the OCR performance.

MS-SSIM~\cite{wang2003multiscale} evaluates the quality of the reconstructed image from various aspects, such as the visual system perceptual ability of the observer. Therefore, compared with SSIM, MS-SSIM can provide a good approximation to perceived image quality and is more suitable for the illumination correction results.

ED~\cite{levenshtein1966binary} is used to evaluate the similarity of two strings. When we fix one string and convert another string to it, the operation we use may include insertion, deletion, and substitution. Each operation will increase the value of ED. Therefore, the higher the ED, the greater difference between the two strings.

CER is also a common metric to evaluate the results of OCR. The CER can be calculated as: $CER=(s+d+i)/(s+d+c)$, where $s,d,i$ are the substitution, deletion and insertion numbers, respectively. And $c$ denotes the number of correct characters. Usually, the lower the CER is, the better the performance will be.

\subsection{Implementation Details}

We implement our method by Pytorch 1.8. For OCR evaluation, we use the latest OCR version (v5.0.1) and pytesseract (v0.3.9). The experiment can be divided into two steps. The first step is to train LPNet and the second step is to train UDoc-GAN. Both processes are carried out on an NVIDIA RTX 2080Ti GPU.

\textbf{LPNet.} As mentioned above, we train LPNet on a modified DocProj dataset. We randomly crop the images into $256 \times 256$ patches. The total training epochs and mini-batch size are set to 200 and 16. An Adam optimizer~\cite{kingma2014adam} with an initial learning rate of $1\times 10^{-3}$ is applied to optimize the network, and the learning rate will be linearly decayed to $5\times 10^{-4}$ in the last 100 epochs.

\textbf{UDoc-GAN.} We apply a U-Net~\cite{ronneberger2015u} shape model for the generator. 
During training, we randomly crop the images into $256 \times 256$ patches. The total training epochs and mini-batch size are set to 300 and 16. 
We use an Adam optimizer which is similar to the LPNet. 
For the generator, we set the initial learning rate as $2\times 10^{-4}$ for the first 200 epochs and linearly decayed to zero in the next 100 epochs. But for the discriminator, the learning rate is double of this setting.

\begin{figure*}[th]
	\begin{center}
		\includegraphics[width=0.88\linewidth]{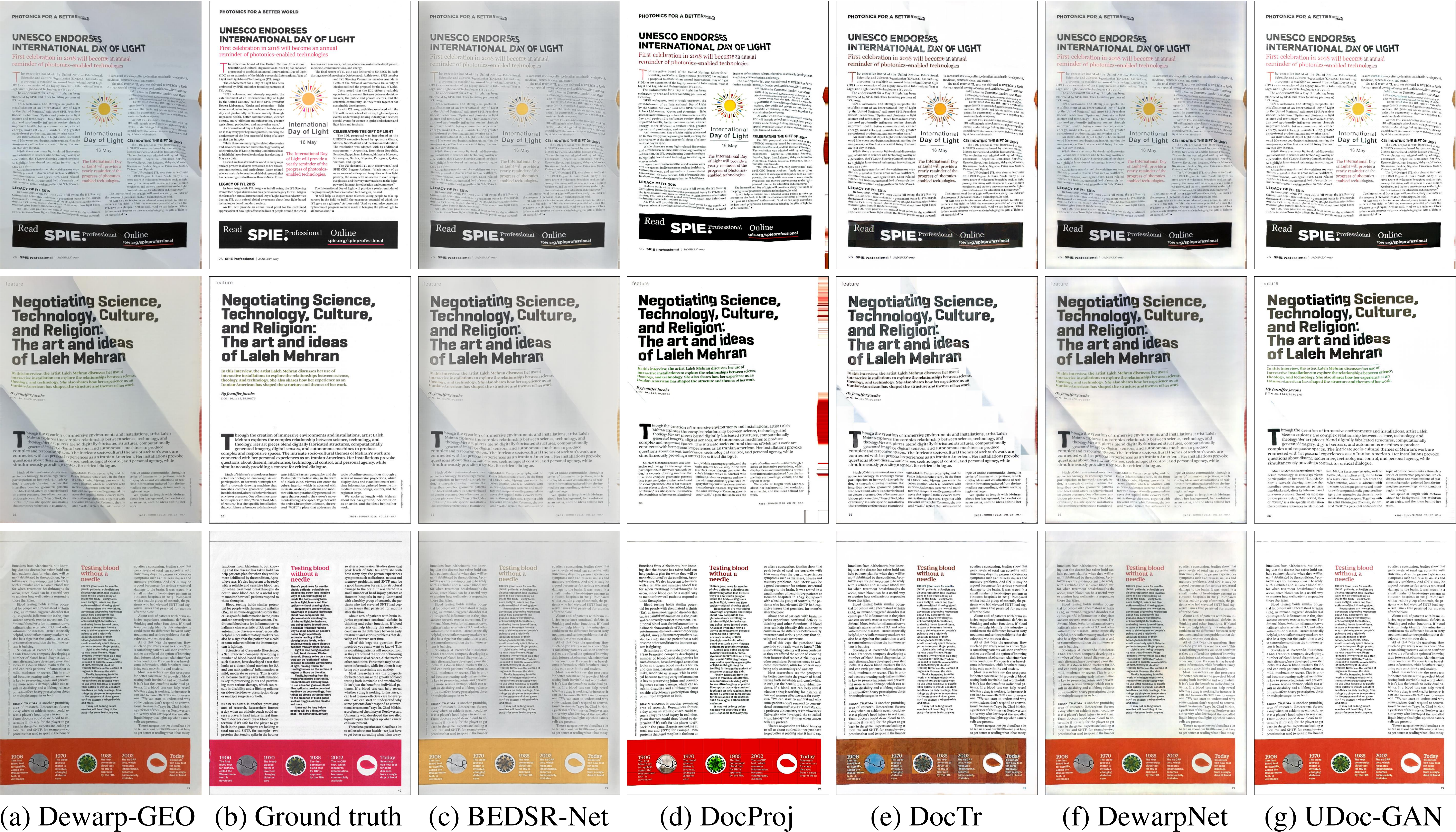}
	\end{center}
	\vspace{-0.1in}
	\caption{Qualitative comparisons between our proposed UDoc-GAN and other state-of-the-art methods. The illumination correction is performed based on the geometric rectification results of DewarpNet.}
	\label{fig:dewarp}
\end{figure*}

\subsection{Experimental Results}

We test our method on the DocUNet Benchmark~\cite{ma2018docunet} after geometric correction of DocTr~\cite{feng2021doctr} and DewarpNet~\cite{9010747}, respectively. We compare a serious of advanced methods~\cite{9156786,feng2021doctr,9010747,li2019document} by quantitative and qualitative results to prove the effectiveness of our method.

\setlength{\tabcolsep}{7pt}
\begin{table}[t]
    \caption{Comparison with all previous works on DocUNet benchmark. DocTr-GEO denotes that the benchmark is geometrically corrected by DocTr. ``$\uparrow$'' indicates the higher the better and ``$\downarrow$'' means the opposite.}
    \centering
    \begin{tabular}{l|ccc}  
    \Xhline{2.5\arrayrulewidth}
    	\textbf{Method} & \textbf{MS-SSIM} $\uparrow$  &\textbf{ED} $\downarrow$  &\textbf{CER} $\downarrow$ \\
    	\hline\hline
    	Distorted Image                & -      & 1789.19 & 0.51  \\
        DocTr-GEO~\cite{feng2021doctr} & -      & 699.34  & 0.20  \\
        BEDSR-Net~\cite{9156786}       & 0.48   & 592.10  & 0.18  \\
        DocProj~\cite{li2019document}  & 0.39   & 533.19  & 0.15  \\
        DocTr~\cite{feng2021doctr}     & 0.48   & 423.00  & 0.13  \\
        \hline
    	UDoc-GAN                       &  \textbf{0.50} & \textbf{396.17} & \textbf{0.12}  \\    	
    \Xhline{2.5\arrayrulewidth}
    \end{tabular}
    \\[6pt]

    \label{tab:doctr}
\vspace{-0.25in}
\end{table}

\textbf{Test on DocTr Geometric Results.} DocTr~\cite{feng2021doctr} geometric correction is one of the state-of-the-art geometric correction schemes, so the result of it is very suitable to be used as a preprocessed result in our illumination correction experiments. We apply all previous state-of-the-art methods to conduct illumination correction. Note that some methods include both geometric correction and illumination correction networks. Here, we only conduct their illumination correction procedures and make comparisons. For DocProj~\cite{li2019document} and DocTr~\cite{feng2021doctr}, we directly execute their public code to produce normal illumination images. However, the code of BEDSR-Net~\cite{9156786} is not publicly available, so we try our best to realize this method, in order to achieve the original results according to the parameters described in this paper, and then apply the corresponding code in our experiment. All these methods are trained on the paired dataset. As shown in Tab.~\ref{tab:doctr}, even if we use a dataset that is unpaired, we yet obtain a performance improvement of CER by 3\% and ED by 7.69\% over the state-of-the-art result by DocTr, and for MS-SSIM, we also achieve the best performance. These results jointly validate the effectiveness of our method. For qualitative results, as shown in Fig.~\ref{fig:doctr}, our method obtains excellent perceptual effects in both text and pictures. Particularly, for the regions containing various color information, our method can still restore the authentic texture, while others have a considerable color deviation. 
We analyze that the light prior and adversarial learning jointly guide the network to learn the difference and relationship between the background light source and the intrinsic statistics of the document.

\setlength{\tabcolsep}{7pt}
\begin{table}[t]
    \caption{Comparison with all previous works on DocUNet benchmark. DewarpNet-GEO denotes that the benchmark is geometrically corrected by DewarpNet. ``$\uparrow$'' indicates the higher the better and ``$\downarrow$'' means the opposite.}
    \centering
    \begin{tabular}{l|ccc}  
    \Xhline{2.5\arrayrulewidth}
    	\textbf{Method} & \textbf{MS-SSIM} $\uparrow$  &\textbf{ED} $\downarrow$  &\textbf{CER} $\downarrow$ \\
    	\hline\hline
    	Distorted Image                & -      & 1789.19 & 0.51  \\
        DewarpNet-GEO~\cite{9010747}   & -      & 792.48  & 0.24  \\
        BEDSR-Net~\cite{9156786}       & 0.45   & 702.93  & 0.22  \\
        DocProj~\cite{li2019document}  & 0.36   & 649.69  & 0.19  \\
        DocTr~\cite{feng2021doctr}     & 0.45   & 576.23  & 0.19  \\
        DewarpNet~\cite{9010747}       & \textbf{0.47} & 756.11  & 0.22  \\
        \hline
    	UDoc-GAN                       & 0.46 & \textbf{558.61} & \textbf{0.18}  \\    	
    \Xhline{2.5\arrayrulewidth}
    \end{tabular}
    \\[6pt]

    \label{tab:dewarp}
\vspace{-0.25in}
\end{table}

\textbf{Test on DewarpNet Geometric Results.} Additionally, we evaluate our methods on another DocUNet Benchmark, which is geometrically corrected by DewarpNet~\cite{9010747}. The illumination correction results of DewarpNet are released, so we directly test it in the same environment. As shown in Tab.~\ref{tab:dewarp}, the quantitative results demonstrate that our method achieves the best performance on ED as well as CER. And for MS-SSIM, our method also has a competitive edge even if other methods are trained on the paired dataset. It should be noted that the geometric correction result of DewarpNet is slightly worse than DocTr, which further proves the generalization capability of our method. The qualitative results are shown in Fig.~\ref{fig:dewarp}, revealing some challenges (\emph{i.e.}, creases, and heterogeneous surfaces). In addition to preserving the color texture, our method can also remove most of the creases to better maintain the overall effect of the document images.

\textbf{Computational efficiency.} In terms of computational efficiency, we compare the inference time and parameter numbers with previous methods. For a fair comparison, we take the settings recommended by~\cite{feng2021doctr}, which employ an NVIDIA GTX 1080Ti GPU, and test on a 1080P resolution image. The overall comparison is shown in Tab.~\ref{tab:time}, even if UDoc-GAN has 29.2M parameters, it takes only 0.08 seconds to process an image on a GPU, which is far less than 3.84 seconds for DocProj and 2.79 seconds for DocTr. This is because, for DocProj, to ensure the quality of the image, the preprocessing of the bilateral mean is required during inference, and for DocTr, the transformer-based method can not directly process high-resolution images. Therefore, the two methods both perform crop operation. To reduce the artifact between patches, there is a one-eighth overlap between patches, which further increases the inference time. In contrast, our method can directly process images without any other redundant operations, which saves a lot of time.

\setlength{\tabcolsep}{9pt}
\begin{table}[t]
    \caption{Comparison of running time and parameters with previous methods.}
    \centering
    \begin{tabular}{l|ccc}  
    \Xhline{2.5\arrayrulewidth}
    	\textbf{Method} & \textbf{Time(s)}  &\textbf{Parameters(M)} \\  
    	\hline\hline
        DocProj~\cite{li2019document}  & 3.84          & \textbf{0.5}          \\
        DocTr~\cite{feng2021doctr}     & 2.79          & 11.6                  \\
        \hline
    	UDoc-GAN                       & \textbf{0.08} & 29.2                  \\    	
    \Xhline{2.5\arrayrulewidth}
    \end{tabular}
    \\[6pt]
    \label{tab:time}
\vspace{-0.25in}
\end{table}

\begin{figure}[t]
	\begin{center}
		\includegraphics[width=0.88\linewidth]{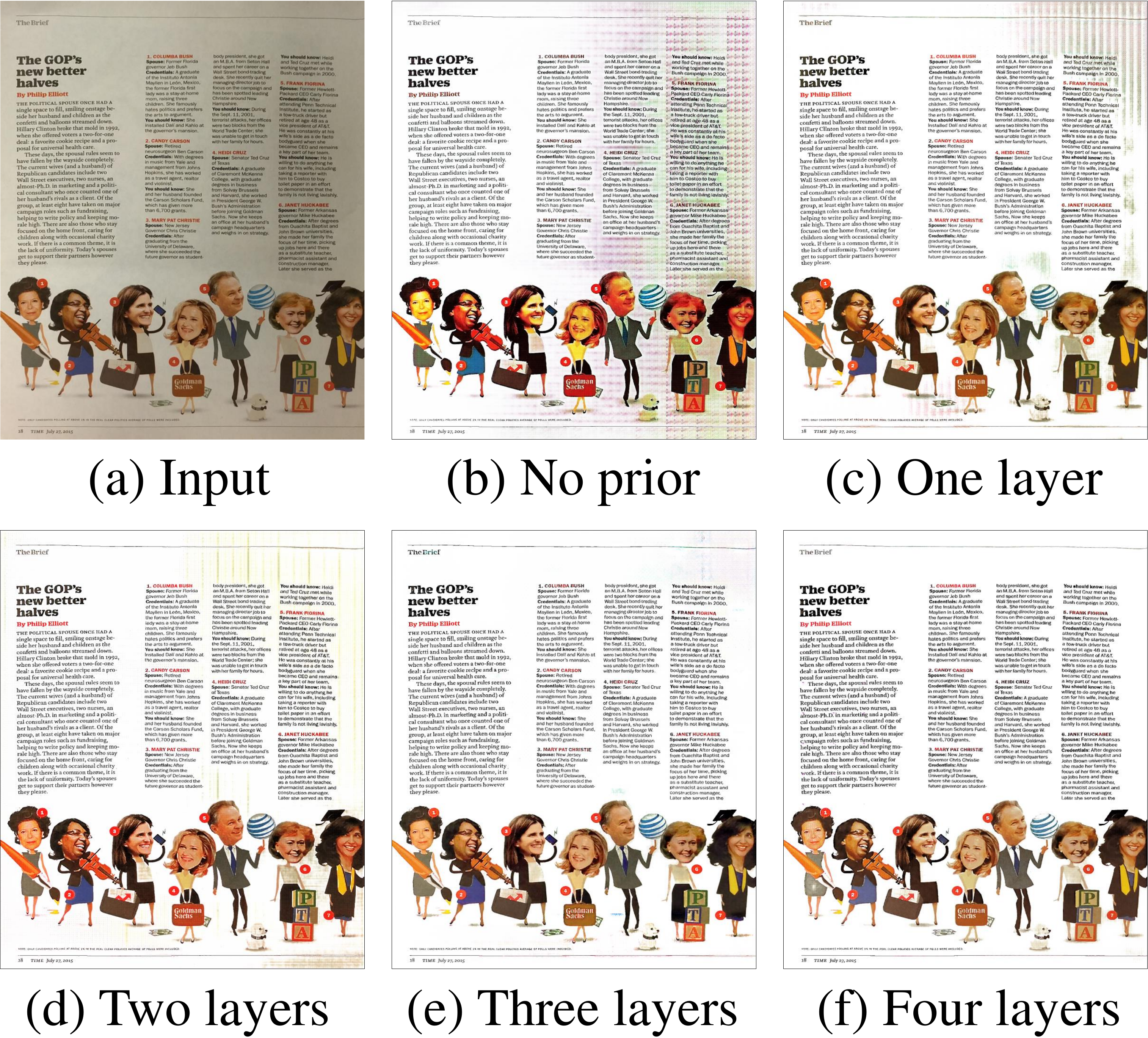}
	\end{center}
	\vspace{-0.1in}
	\caption{Qualitative results of light prior. Our method uses a light prediction network with four Conv layers.}
	\label{fig:prior}
\end{figure}

\subsection{Ablation Study}

In this section, we perform ablation studies to verify the effectiveness of the light prior. Moreover, another ablation study about our full loss functions are conducted to confirm the role of each loss in the experiment. All the ablation experiments are carried out on DocTr~\cite{feng2021doctr} geometric correction results.

\setlength{\tabcolsep}{7pt}
\begin{table}[t]
    \caption{Ablative experiments on the effects of light prior. We also verify the performance of the network when the Conv layer number of LPNet is less. ``$\uparrow$'' indicates the higher the better and ``$\downarrow$'' means the opposite.}
    \centering
    \begin{tabular}{l|ccc}  
    \Xhline{2.5\arrayrulewidth}
    	\textbf{Configuration} & \textbf{MS-SSIM} $\uparrow$  &\textbf{ED} $\downarrow$  &\textbf{CER} $\downarrow$ \\
    	\hline\hline
        No prior               & 0.48 & 605.41  & 0.18  \\
        One layer              & 0.48 & 527.01  & 0.17  \\
        Two layers             & 0.49 & 488.17  & 0.14  \\
        Three layers           & 0.49 & 431.05  & 0.13  \\
        \hline
        Four layers            & \textbf{0.50} & \textbf{396.17} & \textbf{0.12}  \\    	
    \Xhline{2.5\arrayrulewidth}
    \end{tabular}
    \\[6pt]

    \label{tab:prior}
\end{table}

\textbf{Light prior.} We compare the results of using light prior or not. Note that our light prediction network has four Conv layers, and to verify the reliability of the light prior, a comparative experiment is also designed. 
As shown in Tab.~\ref{tab:prior}, the method of not using light prior is worse than any other results. 
Additionally, with the increase in the number of Conv layers, the performance is getting better. 
This process can be regarded as the light prior is becoming more and more reliable.
However, the results are not further improved than the four Conv layers when more Conv layers are applied.
Fig.~\ref{fig:prior} shows qualitative results. 
Without the light prior, the final results exhibit some artifacts in the process of illumination correction.

\begin{figure}[t]
	\begin{center}
		\includegraphics[width=0.88\linewidth]{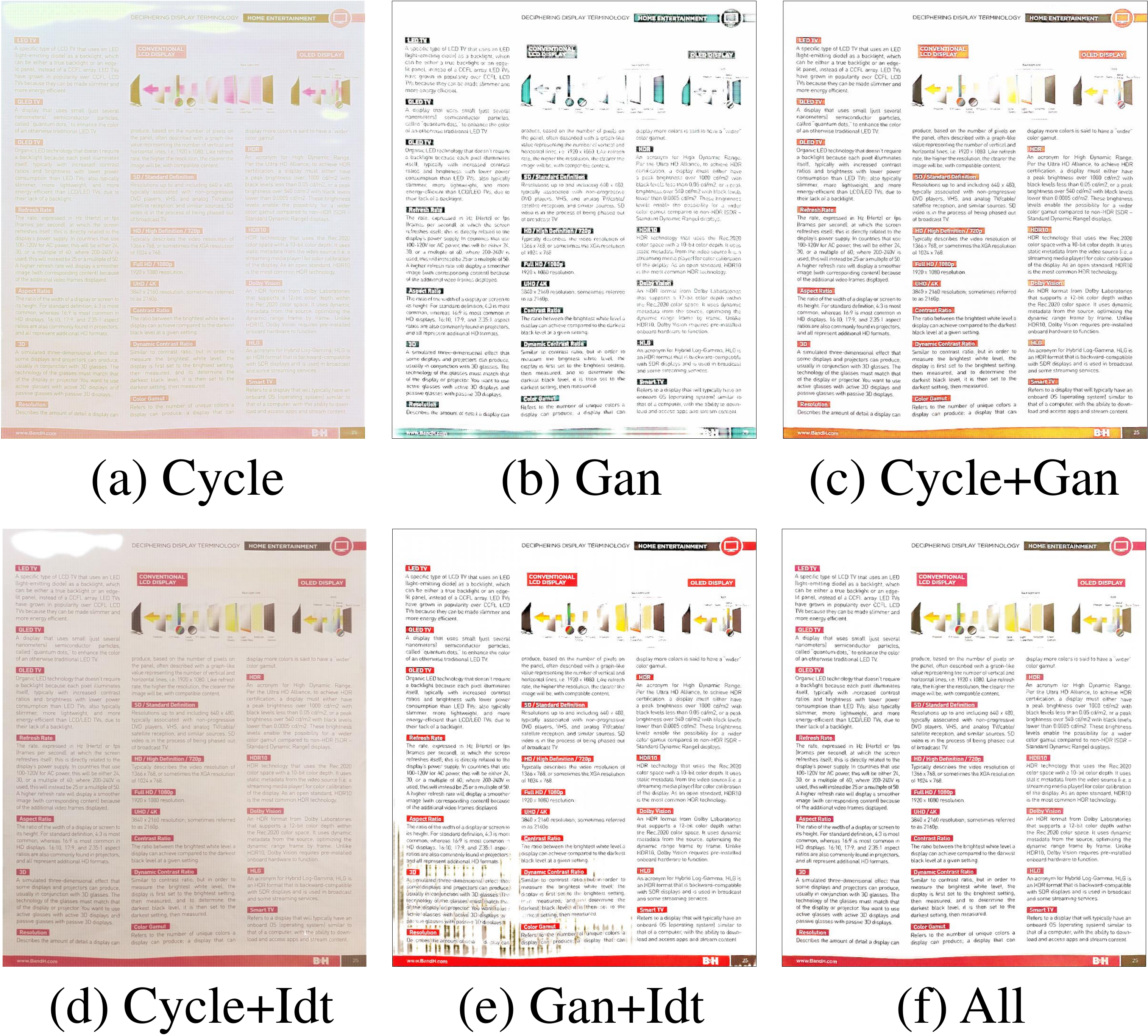}
	\end{center}
	\vspace{-0.1in}
	\caption{Qualitative results of different loss combinations.}
	\label{fig:loss}
\vspace{-0.15in}
\end{figure}

\setlength{\tabcolsep}{5pt}
\begin{table}[t]
	\caption{Ablative experiments on the effects of losses. ``$\uparrow$'' indicates the higher the better and ``$\downarrow$'' means the opposite.}
    \small
	\centering
	\begin{tabular}{l|c|c|c|c|c|c}
		\Xhline{2.5\arrayrulewidth}
		\textbf{loss} & \textbf{(a)} & \textbf{(b)} & \textbf{(c)} & \textbf{(d)} & \textbf{(e)} & \textbf{(f)}  \\
		\hline		
		\hline
		Cycle loss    &$\checkmark$ &              & $\checkmark$ &             &$\checkmark$ &$\checkmark$      \\
		GAN loss      &             & $\checkmark$ &              &$\checkmark$ &$\checkmark$ &$\checkmark$      \\
		Identity loss &             &              & $\checkmark$ &$\checkmark$ &             &$\checkmark$      \\
		\hline
		MS-SSIM $\uparrow$ & 0.48    & 0.50   & 0.49    & 0.48    & 0.50    & \textbf{0.50}    \\
		\hline
		ED $\downarrow$    & 890.65  & 617.39 & 874.42  & 543.59  & 409.11  & \textbf{396.17}  \\
		\hline
		CER $\downarrow$   & 0.30    & 0.18   & 0.26    & 0.17    & 0.13    & \textbf{0.12}    \\
		\Xhline{2.5\arrayrulewidth}
	\end{tabular}
	\\[4pt]

	\label{tab:loss}
\vspace{-0.25in}
\end{table}

\textbf{Loss function.} In Tab.~\ref{tab:loss}, we perform ablation experiments on all loss functions to verify the effect of each loss. We conclude that each term is critical to the results. As shown in Fig.~\ref{fig:loss}(a)(b), we argue that the main function of cycle loss is to maintain the semantic information in images and preserve the one-to-one mapping. But this loss can not guarantee the style features of the images (\emph{i.e.}, it has difficulty distinguishing between normal and abnormal illumination domains). Gan loss can solve this problem and it can produce documents that have more realistic details. Therefore, by combining the two losses, we can obtain relatively ideal images in Fig.~\ref{fig:loss}(c). However, without identity loss, the color threshold may shift. In Fig.~\ref{fig:loss}(c)-(f), we can conclude that the lack of any loss function will affect the results to different degrees.

\section{Conclusion}
In this work, we present a novel generative adversarial framework, named UDoc-GAN, which is the first network using light prior to address the document illumination problem with unpaired data. 
The key idea of our method is to transform the uncertain normal-to-abnormal image translation into a deterministic image translation with the guidance of different levels of ambient light, which can be learned from the illumination image. 
In this way, we realize the translation between normal and abnormal illumination domains.
Extensive experiments with both quantitative and qualitative results validate the effectiveness of our method.

Possible future improvements of this work are as follows: Since the network needs high accuracy of light prior, it is meaningful to construct a larger dataset that contains more diverse illumination conditions, so as to further enhance the recovery capability of tackling various illumination. In addition, removing the uneven illumination effect caused by creases in real document images is a critical issue for the illumination correction task. Therefore, in the future, we intend to predict the location of creases and obtain better correction results by eliminating these creases.

\begin{acks}
This work was supported by the National Natural Science Foundation of China under Contract 61836011 and 62021001. It was also supported by the GPU cluster built by MCC Lab of Information Science and Technology Institution, USTC.
\end{acks}

\bibliographystyle{ACM-Reference-Format}
\balance 
\bibliography{sample-base}

\end{document}